\documentclass[twoside]{article}

\usepackage[accepted]{aistats2022}
\usepackage{hyperref}
\usepackage{url}

\usepackage{amsmath,amsfonts,bm}

\def\eqref#1{equation~\ref{#1}}

\def\1{\bm{1}}

\def\rvx{{\mathbf{x}}}

\DeclareMathAlphabet{\mathsfit}{\encodingdefault}{\sfdefault}{m}{sl}
\SetMathAlphabet{\mathsfit}{bold}{\encodingdefault}{\sfdefault}{bx}{n}

\def\gD{{\mathcal{D}}}

\def\gS{{\mathcal{S}}}

\def\gX{{\mathcal{X}}}

\def\gZ{{\mathcal{Z}}}

\newcommand{\E}{\mathbb{E}}

\DeclareMathOperator*{\argmax}{arg\,max}
\DeclareMathOperator*{\argmin}{arg\,min}

\usepackage{amssymb}
\usepackage{amsmath}
\usepackage{booktabs}
\usepackage{microtype}
\usepackage{subfigure}
\usepackage{graphicx} 
\usepackage{caption}
\usepackage{booktabs} 
\usepackage{wrapfig}
\usepackage{cutwin}
\usepackage{float}
\usepackage{enumitem}
\usepackage{cclicenses}
\usepackage{makecell}
\usepackage{lscape,array}
\newcolumntype{C}[1]{>{\centering\arraybackslash}p{#1}} 
\usepackage[thin, , thinc]{esdiff}
\usepackage{framed}  
\usepackage[font=small,skip=0pt]{caption}
\usepackage{algorithm}
\usepackage{algpseudocode}
\usepackage{titlesec}
\titlespacing*{\section}{0pt}{1.0\baselineskip}{-0.1\baselineskip}
\titlespacing*{\subsection}{0pt}{1.0\baselineskip}{-0.1\baselineskip}
\titlespacing*{\subsubsection}{0pt}{0.5\baselineskip}{-0.1\baselineskip}
\usepackage[round]{natbib}

\usepackage{xcolor}

\newcommand{\rjen}[1]{{\color{blue}RJ: #1}}
\newcommand{\gd}[1]{{\color{green}GD: #1}}
\newcommand{\seti}[1]{{\color{teal}SA: #1}}
\usepackage{accents}
\newlength{\dhatheight}

\begin{document}

%
\runningtitle{Predicting the utility of search spaces for black-box optimization}

%
\runningauthor{Ariafar et al.}

\twocolumn[

\aistatstitle{Predicting the utility of search spaces for black-box optimization:\\a simple, budget-aware approach}

\aistatsauthor{ Setareh Ariafar \And Justin Gilmer\And Zachary Nado }
\aistatsaddress{ Google Research \And   Google Research  \And  Google Research}
\aistatsauthor{ Jasper Snoek \And Rodolphe Jenatton \And George E. Dahl }
\aistatsaddress{ Google Research \And   Google Research  \And  Google Research}]

\begin{abstract}
\vspace{-1em}
Black box optimization requires specifying a search space to explore for solutions, e.g. a $d$-dimensional compact space, and this choice is critical for getting the best results at a reasonable budget. Unfortunately, determining a high quality search space can be challenging in many applications. 
For example, when tuning hyperparameters for machine learning pipelines on a new problem given a limited budget, one must strike a balance between excluding potentially promising regions and keeping the search space small enough to be tractable. The goal of this work is to motivate---through example applications in tuning deep neural networks---the problem of predicting the quality of search spaces conditioned on budgets, as well as to provide a simple scoring method based on a utility function applied to a probabilistic response surface model, similar to Bayesian optimization. We show that the method we present can compute meaningful budget-conditional scores in a variety of situations. We also provide experimental evidence that accurate scores can be useful in constructing and pruning search spaces. Ultimately, we believe scoring search spaces should become standard practice in the experimental workflow for deep learning.
\end{abstract}

\section{Introduction and Motivation}\label{sec:intro}

Solving a black box optimization problem requires defining a search space to optimize over, but in important applications of black box optimization there is not an obvious best search space \emph{a priori}. Selecting the hyperparameters\footnote{Although ``hyperparameter'' has a precise meaning in Bayesian statistics, it is often used informally to describe any configuration parameter of a ML pipeline. We adopt this usage here and depend on context to disambiguate.} of machine learning (ML) pipelines serves as our exemplar for this sort of application, since the choice of search space can be critical for getting good results. If the search space is too broad, the search may fail to find a good solution, but if it is too narrow, then it may not contain the best points. To complicate matters more, the quality of a search space will also depend on the compute budget used to search, with larger budgets tending to favor broader search spaces.


The necessity of selecting good search spaces for black box optimization methods (e.g. Bayesian optimization) presents a barrier to fully automating hyperparameter tuning for ML, and presents challenges for the reproducibility of ML research. Accurately measuring progress in modern ML techniques is notoriously difficult as most new methods either present new hyperparameters to be tuned, or interact strongly with existing hyperparameters (e.g. the optimizer learning rate). As a result, the performance of any new method may depend more on the effects of implicit or explicit tuning rather than on any specific advantage it might hold. Worryingly, methods with initial promising results are often later found to offer no benefits over a baseline that was rigorously tuned in retrospect~\citep{melis2018, merity2018, lucic2018, narang2021}.  In fact, recently it has become commonplace for contradictory results in the literature to be due almost entirely to differing tuning protocols and search spaces~\citep{pineau2020, NadoEtAl2021_largeBatchReality, choi2019empirical}. Ideally, methods papers could provide recommended search spaces for any hyperparameters, but providing good recommendations is challenging without a way to select search spaces for different budgets.

When tuning ML hyperparameters, a search space typically corresponds to a set of hyperparameter dimensions (e.g. the parameters of the learning rate schedule, other optimizer parameters such as momentum, regularization parameters, etc.) and allowed ranges for each dimension. For modern neural networks, it should be possible to define, \emph{a priori}, a maximum conceivable range for any given hyperparameter and whether it should be tuned on a log or linear scale. However, in general, researchers may not know exactly which hyperparameters should even be tuned, and will also have to guess appropriate ranges. Since training neural networks can be so expensive, when faced with small budgets, one should be cautious about adding dimensions or increasing the search space volume, whereas with larger budgets, broader, higher-dimensional search spaces might yield dramatically better results. 

In order to address these issues, we argue that the community should consider the problem of creating \emph{search space scoring methods} that predict the utility of conducting a search (with some black box optimization algorithm), in a particular search space, given a particular budget. Such a score could be informed by initial function observations from a broad, enclosing search space of the sort that can be specified
easily
\emph{a priori}.  A useful score should be able to accurately rank the quality of search spaces in a way that meaningfully depends on the budget used for the search and correlates with the outcomes of a typical search.

Predictive search space scores would have several applications in ML hyperparameter tuning workflows. First and foremost, accurate scores would lead to better decisions when tuning hyperparameters. Practitioners often find themselves interacting with tools such as Bayesian optimization or random search in an iterative loop, tweaking and refining the search space over multiple rounds of experiments until they find a suitable space. Anecdotally, this workflow has become especially common in recent years as larger budgets with higher degrees of parallelism have become available. A search space score could select the most promising and budget-appropriate of several candidate search spaces to try next, help prune bad regions of an initial broad search space, or help match the budget to the space.

As a second application, we can use search space scores to answer scientific questions about experimental results. Although tempting, it does not follow that just because the best point returned by a black box optimization method sets some hyperparameter to a particular value (e.g. uses a $\tanh$ activation function instead of ReLU) other values might not produce just as good results. To answer such a question, one would need to tune the various other hyperparameters for networks assuming a ReLU activation function, separately re-tune them assuming a $\tanh$ activation function, and then compare the results. Alternatively, the same question can be answered by comparing the scores of two search spaces, one that fixed the activation function to $\tanh$ and one that fixed the activation function to ReLU. Budget-conditional scores also make it explicit that the answer depends on the tuning budget, since we might have a situation where, say, the $\tanh$ activation function performs just as well as ReLU when we can afford to carefully tune the learning rate schedule, but when that isn't possible, perhaps ReLU does better. 

Finally, as we alluded to above, accurate, budget-conditional search space scores could be a useful tool to improve the reproducibility of ML research. Scores could be used to recommend search spaces given different budgets for the hyperparameters of a new method, removing some of the guesswork when trying to tune a ML method on a new problem or dataset.

In this paper, we develop a simple method for computing budget-conditional search space scores based on some initial seed data. Taking inspiration from Bayesian optimization, we define a general scoring method based on a utility function applied to a probabilistic response surface model conditioned on the seed data. Our method is agnostic to the specific probabilistic model and thus can make use of future improvements in modeling for Bayesian optimization. Specifically, in this work, we:

\begin{itemize}[noitemsep,topsep=0pt,parsep=0pt,partopsep=0pt]
    \item Motivate the search space scoring problem for machine learning hyperparameter tuning and argue it is worthy of more study.
    \item Introduce a simple method for computing budget-conditional scores, show that it can be accurate in experiments on real neural network tuning problems, and demonstrate its failure modes.
    \item Provide empirical evidence that these scores could be useful in many tuning workflows.
    \item Present examples of how one might answer experimental questions using search space scores.
\end{itemize}


\section{Related Work}
Bayesian optimization has a deep literature with well established methodologies for experimental design~\citep{shahriari2016, garnett_bayesoptbook_2022}. It has become the state-of-the-art methodology for the tuning of hyperparameters of ML algorithms~\citep{bergstra2011, snoek2012, turner2021a}.  While the literature is full of methodological improvements to Bayesian optimization, 
they generally assume a reasonably well chosen search space.

Although we do not create a new Bayesian optimization method in this work so much as present a different problem formulation, methods that focus on performing Bayesian optimization on very large search spaces or identifying promising local regions within a larger space are related. These methods do, in some sense, dynamically refine the search space or switch between smaller spaces.
\citet{wang2016} and~\citet{letham2020re} perform a random projection into a smaller latent space to enable optimization on very large search spaces under the assumption that many dimensions are either strongly correlated or not useful.  \citet{eriksson2019scalable} introduce TuRBO to optimize over large scale, high dimensional problems using a bandit-style algorithm to select between multiple local models.  \citet{mcleod2018optimization} developed a method called BLOSSOM, which switches between a global Bayesian optimization and more traditional local optimization methods.  \citet{assael2014} divide up the search space into several smaller search spaces in the context of a treed Gaussian process, so that local GPs are well-specified on each of the smaller regions.  \citet{wang2020learning} partition a search-space by means of Monte Carlo Tree Search.

Search-space \textit{pruning} strategies~\citep{wistuba2015hyperparameter}, using sparsity-inducing tools~\citep{cho2019reducing}, or search-space \textit{expansion} approaches~\citep{shahriari2016unbounded, nguyen2019filtering} dynamically adjust the search space during optimization.
Moreover, when previous data about related tasks is available, \citet{perrone2019learning} prune search spaces using transfer learning. Through theoretical analysis,~\citet{tran2020sub} develop a method to dynamically expand search spaces while incurring sub-linear regret.  The idea of transforming the search space over the course of the optimization also appears within the context of adjustments made by developers of an ML system~\citep{stoll2020}.
Constrained Bayesian optimization~\citep{gelbart2014, miguel2014, gardner2014, gelbart2016} provides a framework for modeling infeasible regions of the search space through the framework of constraints.  

Although several technically sophisticated and effective methods exist to operate on search spaces that are too small, too broad or ill-behaved, we are not aware of work that accomplishes our goal, namely developing a tool to score search spaces, primarily within the context of ML hyperparameter tuning.



\section{Scoring search spaces given budgets}
Our goal is to score a search space $\gS$, 
conditional on a budget $b$, such that drawing $b$ points from a higher ranking search space is likely to lead to better (lower) function values from some function $f(x)$.
In other words, our scores should reflect whether lower values of $f$ reside inside a search space and how easy it will be, on average, to find them with only $b$ queries.
In order to make this notion more precise, the following notation will be useful.
Given a search space $\mathcal{S} \subseteq \mathcal{X}$, we denote a uniform sample from the set $\mathcal{S}$ with $x \sim U(\mathcal{S})$, and use $\mathbf{x} \sim U_b(\mathcal{S})$ to denote a uniform i.i.d.~batch of $b$ samples. Now we can write the expected minimal function value found after sampling $b$ points from $\mathcal{S}$ as
\vspace{-0.5em}
\begin{equation*}
    \alpha(\gS, b) = \E_{\rvx \sim U_b(\gS)} \left[\min_{x \in \rvx} f(x)\right].
\end{equation*}
Then, given two search spaces $\gS_1$ and $\gS_2$ together with a query budget $b$, $\alpha(\gS_1, b) < \alpha(\gS_2, b)$ would indicate that $\gS_2$ should be preferred to $\gS_1$ at budget $b$.





Unfortunately, there are two immediate challenges with this approach. First, the true function $f$ is not known, and second, we can only access $f$ through expensive, potentially noisy, evaluations. Therefore, if we assume access to some observed data $\gD=\{(x_j,y_j)\}_{j=1}^n$ where $\forall j ~x_j \in \gX $ and $y_j = f(x_j)+\varepsilon=\Tilde{f}(x_j)$, with $\varepsilon$  denoting the observation noise, 
we can construct a probabilistic model of noise-corrupted $f$ which we denote by $y = f(x) + \varepsilon \sim p(y | x, \gD)$.
Our goal now is to leverage a learned probabilistic model $p(y | x, \gD)$ to design search space scoring functions $\hat{\alpha}(\cdot, b, p)$ that will be faithful to the original criterion $\alpha(\cdot, b)$. By ``faithful'', we mean that it should preserve the ranking that $\alpha(\cdot, b)$ induces, i.e.
$
    \alpha(\gS_1, b) < \alpha(\gS_2, b)
    \ \ \Rightarrow\ \ 
    \hat{\alpha}(\gS_1, b, p) < \hat{\alpha}(\gS_2, b, p).
$

In most cases, we actually want to improve upon some incumbent, best observed value $y^+$.
In such cases, our scoring functions can target more general \emph{utility functions} $u(y^+, f(\mathbf{x}))$. Similar batch improvement-based utilities exist in the Bayesian optimization literature as a part of batch acquisition functions, which we can directly employ here. For example, $u$ might encode \emph{whether} or \emph{how much} the best $f$ values of the $x$ locations in $\mathbf{x}$ can improve upon $y^+$. However, because we do not have access to $f(\mathbf{x})$, we can instead estimate the expected utility $\E_{\hat{\mathbf{y}} \sim p(\mathbf{y} | \mathbf{x}, \mathcal{D})} \left[ u(y^+,\hat{\mathbf{y}}) \right]$ in place of $u(y^+, f(\mathbf{x}))$. 
If we adopt improvement over $y^+$ as our utility function, our expected utility expression becomes the multi-point expected improvement (b-EI) batch acquisition function~\citep{wang2016parallel}. Similarly, if we use binarized improvement as our utility, our expected utility will reproduce the multi-point probability of improvement (b-PI) batch acquisition function~\citep{wang2016parallel}. Of course unlike with the batch acquisition functions common in Bayesian optimization, we are not trying to find the location of the next batch of query points, but are instead predicting the score of a search space by averaging over batches of points within that search space.

\newpage
\textbf{Definition of the score:} For concreteness, consider the case where we define the utility of a batch as the improvement over some incumbent value $y^+$. 
Ideally, we would directly compute the average batch expected improvement, i.e. the average b-EI, under $p(y | \mathcal{D}, x)$ for random batches of size $b$ sampled from $\mathcal{S}$, namely
\begin{multline} \label{eq:score-def}
    \hat{\alpha}(\gS, b, p, y^+) = \\ 
    \E_{\mathbf{x} \sim U_b(\gS), \hat{\mathbf{y}} \sim p(\mathbf{y} | \mathbf{x}, \mathcal{D})} \left[  \max \left( 0, y^+ - \min(\hat{\mathbf{y}}) \right) \right],
\end{multline}
or equivalently, $\E_{\mathbf{x} \sim U_b(\gS)} \left[\text{b-EI}(\mathbf{x}\rvert y^+) \right].$ In addition to this score, we define an analogous score based on b-PI using binarized improvements.
We refer to these scores as mean-b-EI and mean-b-PI in the remainder of the text.
As a simple alternative, we can also replace the mean with respect to $\rvx \sim U_b(\gS)$ with the median. This change will lead to a slightly modified definition with hopefully less sensitivity to outlier utilities. We call the median-based versions of \eqref{eq:score-def} median-b-EI and median-b-PI (with binarized improvements).


\textbf{Computing search space scores in practice:}
In order to compute any scores in practice, we must first fix a probabilistic model.
In this work, we assume a Gaussian Process (GP) model on $f$ and evaluate the expected utility by taking the expectation of $u$ with respect to 
$p(\mathbf{y}|\mathbf{x},\gD)$, which is a multivariate Gaussian. See \ref{app-gp} and \citep{rasmussen2003gaussian} for more details on the GP structure and inference.
Assuming a GP model, the expected utilities b-PI and b-EI can be calculated using high-dimensional multivariate Gaussian CDFs, however, this is not practical especially for large values of $b$ \citep{frazier2018tutorial}.
As an alternative, we use a simple Monte Carlo approach to estimate the expectations in \eqref{eq:score-def}; we draw samples $\mathbf{x}=\{x_i\}_{i=1}^b$ from $U_b(\gS)$, then form the GP posterior $p(\mathbf{y} | \mathbf{x}, \mathcal{D})$ conditioned on the sampled input locations, and then finally compute the average.
Similarly, we estimate the expectation over the different input locations by sampling a number of different batches of $\mathbf{x}$ and averaging. 
We refer to the result of this computation as the \emph{predicted score} for $\gS$ at budget $b$. In addition to the \emph{ex-ante} predicted score, for the purposes of validation, we can also compute an \emph{ex-post} score based on the empirical utility, that is $\E_{\rvx \sim U_b(\gS)}\left[u(y^+,\Tilde{f}(\rvx))\right]$ using real, noise-corrupted evaluations of $f$ at $\rvx$.
We call this score the \emph{empirical score} for $\gS$ at budget $b$. As we will discuss later, the empirical scores are useful to validate the accuracy of predicted scores retrospectively.

\subsection{Example applications of the scores}\label{example-apps}
Our predicted scores have a variety of potential uses in the deep learning experimental workflow. We present three illustrative examples of how they could provide value, although there will inevitably be other cases too.

\subsubsection{Ranking of search spaces}\label{example:rank-ss}
The simplest application, which we already have alluded to, is to rank a set of search spaces $\mathbb{S}=\{\gS_t\}_{t=1}^T$ that are all subsets of an enclosing loose search space $\gX$ based on their predicted score, at budget $b$, given some initial data $\mathcal{D}$ from $\gX$. Ranking search spaces lets us pick what search space to explore next in workflows that construct multiple search spaces in sequence or can answer questions about results we have already posed as ranking questions. See Algorithm \ref{alg:pred-scores} for details. Note that the initial data $\mathcal{D}=\{(x_j,y_j)\}_{j=1}^{n}$ could be obtained in a variety of ways, for example by uniformly sampling and evaluating $n$ points from $\gX$ or by $n$ rounds of Bayesian optimization on $\gX$.
%
%
%
 Although not useful for the applications we consider, we can also imagine a retrospective version of Algorithm~\ref{alg:pred-scores} that replaces predicted scores with empirical scores that we can use to validate our predicted rankings.
 \vspace{-1em}
\begin{algorithm}[htp!]
\caption{Predicted search space ranking}\label{alg:pred-scores}
\begin{algorithmic}[1]
\Require Budget $b$, search space set $\mathbb{S}$, 
initial data $\mathcal{D}$ 
\State $y^+ \leftarrow$ min objective in $\gD$

\State Fit probabilistic model $p(y | x, \gD)$
\For{$\gS $ in $ \mathbb{S}$:}
        \State $\text{predicted-scores}\left[\gS\right]=\hat{\alpha}(\gS,b,p,y^+)$
\EndFor\\
\Return predicted-scores
\end{algorithmic}
\end{algorithm}
\vspace{-1em}

\vspace{-.1em}
\subsubsection{Pruning bad search space regions}\label{exampl-prune-bad}

With a modest amount of data from an overly broad initial search space $\gX$, it should be possible to quickly remove bad regions of the space using search space scores, leaving a final pruned search space that is more efficient to search than the original. Although exploring more sophisticated pruning algorithms based on these scores would be an interesting avenue for future work, here we present a simple proof-of-concept algorithm that does one round of pruning based only on a single initial sample of observations. Our algorithm assumes access to a subroutine \texttt{propose\_search\_spaces}$(\gX,T)$ that proposes candidate search spaces $\{\gS_t\}_{t=1}^T$, where $\forall t,\gS_t \subset \gX$. We describe different possible strategies to implement such a function in \ref{subsec:app-ss}. The algorithm also makes use of a subroutine \texttt{sample\_observations}$(\gS, \Tilde{f}, b)$ that selects $b$ points in $\gS$ and evaluates them using the objective function $\Tilde{f}$, producing the resulting $(x,y)$ pairs as output. The \texttt{sample\_observations} subroutine could be implemented with uniform random search or a Bayesian optimization algorithm. Given a budget of $B= b_1 + b_2$, Algorithm~\ref{alg:ss-prune} uses the first $b_1$ points as initial exploratory data to prune the starting search space and then samples the remaining $b_2$ points from the pruned space, finally returning the best point.

\vspace{-4em}
\subsubsection{Deciding whether or not to tune a particular hyperparameter}\label{example-hp-importance}
Given infinite resources, tuning any given hyperparameter can never do worse than setting it to a fixed value. However, in practice, when constrained by a finite budget, in some cases fixing certain hyperparameters will outperform tuning them. Although there are many ways to determine the relative importance of different hyperparameters (e.g. by examining learned length scales of an ARD kernel or approaches such as \citet{rijn2017}), determining when to tune or fix specific hyperparameters is a slightly different question that inherently involves reasoning about the available tuning budget. Thankfully, with budget-conditional search space scores, this question is quite straightforward to answer since we can reduce it to a question about ranking search spaces. Specifically, let $\gX^d$ be the $d$th hyperparameter and suppose the search space $\gX$ constrains $\gX^d \in \left[l^d,u^d\right]$. Let $\gZ$ be a copy of search space $\gX$ where $\gX^d=\delta$, for some value $\delta \in \left[l^d,u^d\right]$. To decide whether, given a budget of $b$ evaluations, we can afford to tune $\gX^d \in \left[l^d,u^d\right]$ or we should instead fix $\gX^d = \delta$, we can apply Algorithm~\ref{alg:pred-scores} to score the search spaces $\gX$ and $\gZ$ and see which one obtains the better score. In practice, we might be interested in considering a $\delta$ that is either some standard default value of the hyperparameter or the value it takes on at the best point observed so far.
\begin{algorithm}[htp!]
{\color{black}
\caption{One-shot search space pruning}\label{alg:ss-prune}
\begin{algorithmic}[1]
\Require Budget split $B=b_1+b_2$, objective $\Tilde{f}$, $T$,  \texttt{sample\_observations},
 \texttt{ propose\_search\_spaces}


\State $\gD_1 \gets \texttt{sample\_observations}(\gX, \Tilde{f}, b_1)$
\State $y^+ \leftarrow$ min objective in $\gD_1$
\State Build probabilistic model $p(y | x, \gD_1)$
\State $\{\gS_1,\dots,\gS_T\} \gets \texttt{propose\_search\_spaces}(\gX, T)$
\State $\gS^\star = \argmax_{\gS_t} \hat{\alpha}(\gS_t, b_2, p,y^+)$ 
\State $\gD_2 \gets \texttt{sample\_observations}(\gS^\star, \Tilde{f}, b_2)$\\
\Return The best value $(x_{\min}, y_{\min})$ from $\gD_1 \cup \gD_2$
\end{algorithmic}
}
\end{algorithm}
\vspace{-2em}

\section{Experiments}\label{sec:experiments}
\begin{figure*}[t!]
  \centering
\includegraphics[width=0.76\textwidth]{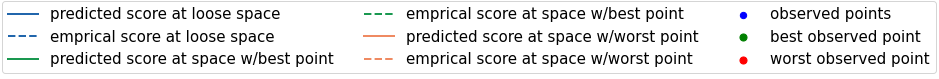}\quad\quad\quad
 \includegraphics[width=0.3\textwidth]{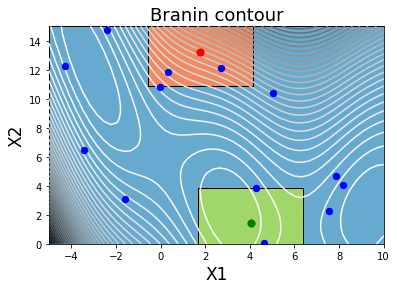}\quad\quad
  \includegraphics[width=0.3\textwidth]{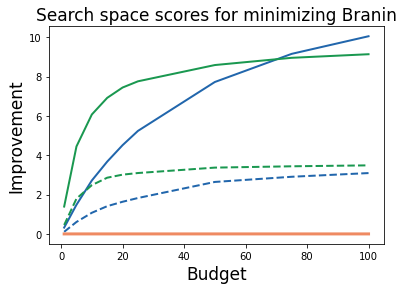}\quad

\caption{\small Scoring Branin function search spaces: given fifteen random observations (left), we compare three search spaces: the enclosing broad search space (blue), and two equal-volume spaces centered around the best and the worst observations (green and orange). We report the predicted mean-b-EI scores (right: solid lines) and compare with empirical scores observed from random sampling true values within each search space (right: dashed lines). The predicted scores correctly show the orange search spaces is worse, and that the gap between the others shrinks at larger budgets.
}
\label{fig:branin-demo}
\vspace{-1em}
\end{figure*}
\begin{figure*}[ht]
  \centering
  \centering
\includegraphics[width=0.7\textwidth]{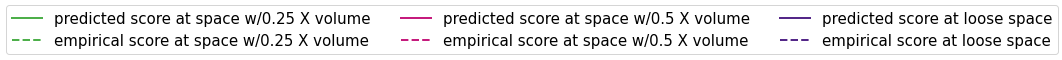}\quad\quad\quad\quad
\includegraphics[width=0.3\textwidth]{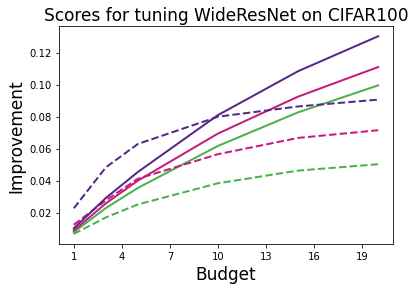}
\includegraphics[width=0.3\textwidth]{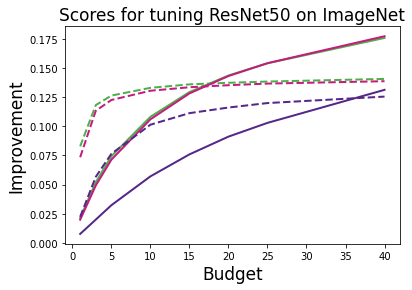}\quad

\caption{\small Comparing search spaces across budgets for tuning a WideResNet on CIFAR100 (left) and a ResNet50 on ImageNet (right). For both, the predicted scores successfully preserve the search space ranking of the empirical scores.
}
\label{fig:ss-scores}
\end{figure*}

In addition to synthetic functions to build intuition, given our focus on machine learning applications, we use several realistic problems from hyperparameter tuning for deep neural networks to demonstrate our search space scoring method. Our experiments generally follow the three example applications in Section~\ref{example-apps}, with the details of the synthetic functions and hyperparameter tuning problems we consider, below.

\subsection{Experiment \& optimization task details}\label{subsec:exp-details}
We instantiate our GPs following the standard design choices of~\citet{eriksson2019scalable}; see \ref{app:gp-details} for details. Initial observations to train the GP are uniformly sampled from the base search space $\gX$. Unless otherwise specified, scores are mean-b-EI and computed with with 1000 sampled x-locations and 1000 samples from the GP posterior at each x-location.
We observed that other variations including mean-b-PI, median-b-EI and median-b-PI produce generally similar results. See \ref{subsec:diff-scores} for more details. All neural network experiments used a polynomial learning rate decay schedule parameterized by an initial learning rate, a power, and the fraction of steps before decay stops; the learning rate reaches a final constant value 1e-3 time the initial learning rate (see \ref{app:ss-details} for details). 
The broad search space for all neural net tuning problems has seven hyperparameters: the initial learning rate $\eta$, one minus momentum $1-\beta$, the decay schedule power $p$, decay step factor $
\tau$, dropout rate $r$, $l_2$ decay factor $\lambda$, and label smoothing $\gamma$. See Table \ref{tab:loose-base-ss} for the scaling and ranges. We collected our neural network tuning data on CIFAR100, ImageNet and LM1B using open source code from \citet{init2winit2021github}.  An implementation of our score function and its applications will be available at \url{https://github.com/anonymous/spaceopt}.




\textbf{Synthetic tasks}: We use two synthetic functions in our experiments, the two-dimensional Branin function and the six-dimensional Hartman function \citep{eggensperger2013towards}. The base search space for Branin is $\gX=[[-5.,10.],[0.,15.]]$. The base search space for the Hartman function is $[0,~1]^6$. For Branin, in addition to the base search space $\gX$, we make use of two additional search spaces, $\gS_1$ and $\gS_2$, that both have roughly $10\%$ of the volume of $\gX$, with $\gS_1$ centered around $x_{\text{best}}=\argmin_{y_j \in \gD}~ y_j$ and $\gS_2$ centered around $x_{\text{worst}}=\argmax_{y_j \in \gD}~ y_j$ (see \ref{subsubsec:app-ss-around-point} for additional search space construction details). 



\textbf{CIFAR100}:
On CIFAR100 \citep{cifarKrizhevsky09learningmultiple} we tune a WideResNet \citep{zagoruyko2016wide} using two hundred training epochs.
In addition to the base search space $\gX$, we make use of two additional search spaces $\gS_1$ and $\gS_2$, where $\gS_1$ is generated following  \ref{subsubsec:app-ss-around-point} around $x_{\text{median}}=\text{arg\,median}_{y_j \in \gD}~ y_j$ with roughly $25\%$ of the volume of $\gX$ and $\gS_2$ is generated similarly around $x_{\text{median}}$ with roughly $50\%$ of the volume of $\gX$.
Note that since both $\gS_1$ and $\gS_2$ are centered around the same point, $\gS_2$ encloses $\gS_1$ and, as always, $\gX$ encloses both. For details, see \ref{app:ss-details}.

\textbf{ImageNet}: On ImageNet \citep{imagenet} we tune ResNet50 \citep{resnet2016}, trained using ninety epochs.
In addition to $\gX$, we make use of two additional search spaces $\gS_1$ and $\gS_2$, where $\gS_1$ is generated following  \ref{subsubsec:app-ss-around-point} centered at $x_{\text{best}}=\argmin_{y_j \in \gD}~ y_j$ with roughly $25\%$ of the volume of $\gX$ and $\gS_2$ is also centered on $x_{\text{best}}$, but with roughly $50\%$ of the volume of $\gX$.
Once again $\gS_1 \subset \gS_2 \subset \gX$. See ~\ref{app:ss-details} for additional details.

\textbf{LM1B}: For LM1B \citep{lm1b} we train Transformer model from \cite{vaswani2017attention}, also referred to as the post-layer-norm Transformer, for 75 epochs.  We tune the same seven hyperparameters as above for CIFAR100 and ImageNet.

\subsection{Ranking search spaces across budgets}\label{subsec:ranking}

When ranking search spaces, our scores should exhibit three qualitative behaviors that are sometimes in tension. 
First, scores should prefer search spaces where $p(y | x, \gD)$ predicts better function values. Second, as the budget increases, scores should be less risk averse. In other words, the score will be less sensitive to how hard good points are to locate \emph{within} the search space, as long as good points exist (according to $p(y | x, \gD)$).
Third, predicted scores should generally agree with the empirical ranking.
Our goal here is to illustrate the different types of behavior the predicted scores can have and show that, although our desiderata are sometimes in conflict, nonetheless on these example problems we see sensible behaviors (we discuss failure modes in Section \ref{sec:fail}). We use GPs conditioned on a modest number of observed points: $\lvert\gD\rvert = 15$ for Branin and tuning WideResNet on CIFAR100, and $\lvert\gD\rvert = 20$ for tuning ResNet50 on ImageNet.

We can see several examples of mean-b-EI preferring search spaces where the GP predicts there are good function values or, equivalently, penalizing search spaces where the GP predicts there are no good function values. On Branin (Fig.~\ref{fig:branin-demo}), $\gS_1$ (green) and $\gS_2$ (orange) are disjoint and $\gS_1$ holds the incumbent $y^+$ whereas $\gS_2$ contains worse observations. And indeed we see $\gS_2$ decisively ranked worse (in agreement with the empirical score ranking). For CIFAR100 (Fig.~\ref{fig:ss-scores}, left), $\gS_1 \subset \gS_2 \subset \gX$, but $\gS_1$ and $\gS_2$ are both centered on a mediocre point and the best point we measured is outside of them both. Presumably the GP posterior assigns a relatively high probability to good points being found outside of $\gS_1$ and $\gS_2$ and thus we see $\gX$ ranked above $\gS_2$ ranked above $\gS_1$, once again agreeing with the empirical score ranking. For ImageNet (Fig.~\ref{fig:ss-scores}, right), $\gS_1 \subset \gS_2 \subset \gX$ and both the inner search spaces are centered on the same very good observation that the GP also conditions on, so the more interesting question is whether the predicted ranking agrees with the empirical score ranking, which it does.
\begin{figure*}[t!]
  \centering
\includegraphics[width=0.385\textwidth]{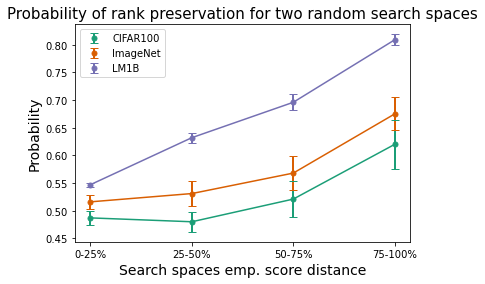}\quad
\includegraphics[width=0.37\textwidth]{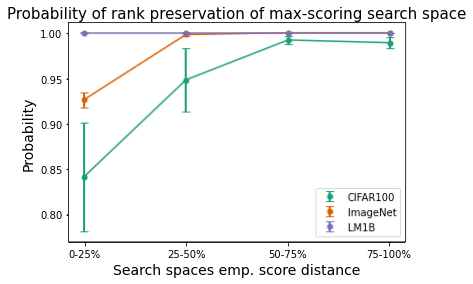}
\caption{\small Probability of preserving the rank of \textbf{(1)} two random search spaces (left) and \textbf{(2)} a random search space and the best (max-scoring) search space (right). We consider three tuning problems including a WideResNet on CIFAR100, a ResNet50 on ImageNet and a Transformer model on LM1B. We report rank accuracy as a function of how close the empirical scores of the search spaces are. The closer the search spaces, the harder their correct ranking will be to preserve. In both cases, we see predicted score rank accuracy improve as the search spaces differ more.}
\label{fig:prob-rank-preserve}
\end{figure*}

\begin{figure*}[t!]
  \centering
\includegraphics[width=0.3\textwidth]{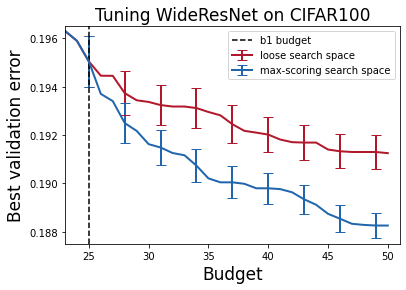}
\includegraphics[width=0.3\textwidth]{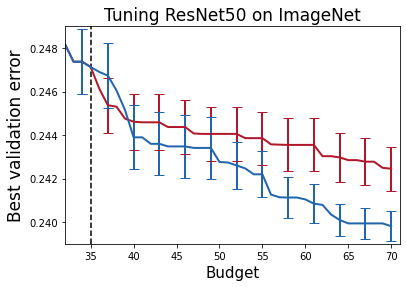}\quad
\caption{\small Performance of random search applied to the loose search space (red) and the max-scoring pruned search space (blue). The score-guided random search method (blue) splits and spends the budget as explained in Alg. \ref{alg:ss-prune}.
The performance of the $b_1$ samples used to train the GP is shown before the vertical line. 
For both WideResNet on CIFAR100 (left) and ResNet50 on ImageNet (right), the pruning provides a clear advantage by the time the budget is exhausted.
}
\label{fig:baseline-vs-boosted}
\end{figure*}
We can also see how the predicted scores become more risk tolerant (care less about how unlikely random search is to find the good points in a search space) in our three example problems. On Branin, $\gS_1 \subset \gX$ and as the budget increases, eventually the predicted score ranks $\gX$ above $\gS_1$, which is reasonable \emph{even though it disagree with the empirical score ranking}. Both $\gX$ and $\gS_1$ contain optimal points, but the GP has no way of knowing $\gS_1$ contains an optimum, so eventually the score prefers the larger, enclosing search space. For CIFAR100, although the outermost search space is always preferred, the gap widens as the budget increases, as we would hope. For ImageNet, the outermost enclosing search space happens to be ranked worse for the budgets we plotted. However, although we don't have enough data to extend the empirical scores to a high enough budget, eventually, around $b=1000$, the predicted score for the $\gX$ actually overtakes the others.

\subsection{Ranking randomly generated search spaces}\label{subsec:rank-preserve}
In \ref{subsec:ranking}, we showed that our scores correctly rank a set of manually defined search spaces. Here, we estimate the probability of preserving the rank of pairs of randomly generated search spaces. We consider three tuning problems including a WideResNet on CIFAR100, a ResNet50 on ImageNet and a Transformer model on LM1B, all scored with $\gD=20$ and budget 15. We generate search spaces from the distribution described in Appendix \ref{subsubsec:app-ss-random} with volume reduction rates $\rho \in \{0.1,~ 0.2,\dots,0.9\}$ with 50 search spaces per rate. 
Then, for two random search spaces $\mathcal{S}_1$ and $\mathcal{S}_2$, we compute the probability of having $\texttt{emp\_score}(\mathcal{S}_1) > \texttt{emp\_score}(\mathcal{S}_2) \Rightarrow \texttt{pred\_score}(\mathcal{S}_1) > \texttt{pred\_score}(\mathcal{S}_2)$. Clearly, the \emph{closer} $\mathcal{S}_1$ and $\mathcal{S}_2$ are in empirical score, the more likely it is for noise in both scores to cause a benign disagreement. Furthermore, the particular distribution of search spaces we consider will induce an arbitrary distribution over the distance between pairs of search spaces. Therefore, we report the probability of correctly ranking as a function of how close $\mathcal{S}_1$ and $\mathcal{S}_2$ are, based on $| \texttt{emp\_score}(\mathcal{S}_1) -\texttt{emp\_score}(\mathcal{S}_2) |$ quantiles (e.g., ``0-25 $\%$.'' contains all the smallest distances until the first quartile). We estimate the probability using 2000 samples, where each sample selects two search spaces at random from the pool of 450 we generated; Fig. \ref{fig:prob-rank-preserve} reports the mean and standard error over 10 runs (each run has a different random $\gD$). We also replicate the experiment for random search spaces vs the max-scoring (according to the empirical score) search space, $\mathcal{S^*}$.

\begin{figure*}[t!]
  \centering
  \includegraphics[width=0.8\textwidth]{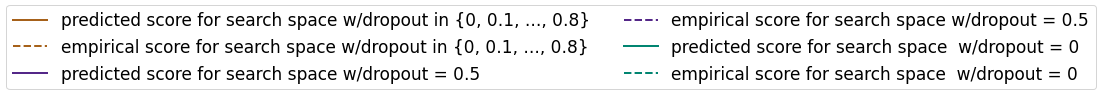}\quad\quad\quad\quad
 \includegraphics[width=0.29\textwidth]{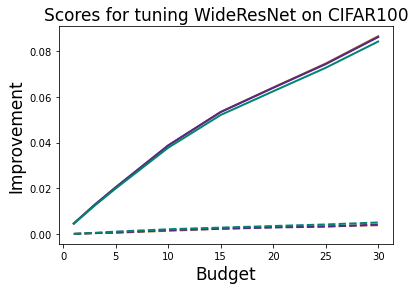}
 \includegraphics[width=0.29\textwidth]{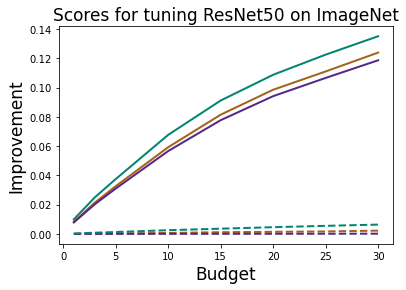}
\includegraphics[width=0.29\textwidth]{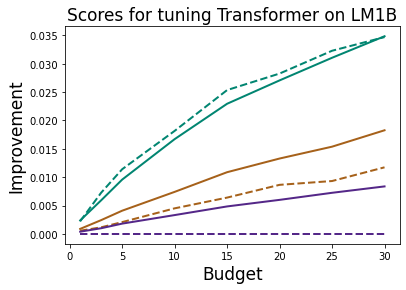}
\caption{\small Deciding whether to tune over dropout, fix it to 0, or fix it to 0.5 for WideResNet on CIFAR100, ResNet50 on ImageNet and Transformer on LM1B. 
}
\label{fig:hp-importance}
\end{figure*}
\begin{figure}[t!]
  \centering
\includegraphics[width=0.45\textwidth]{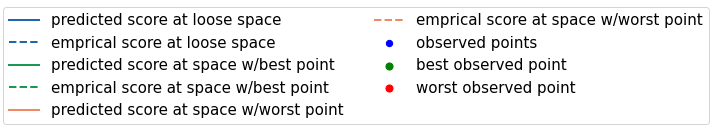}\quad\quad\quad\quad
\includegraphics[width=0.23\textwidth]{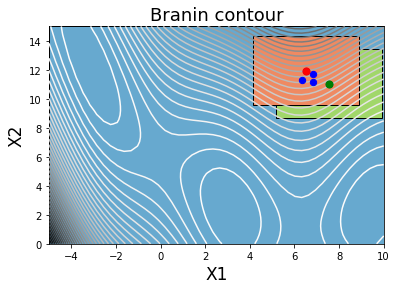}
\includegraphics[width=0.23\textwidth]{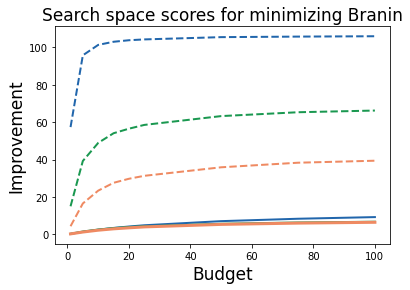}

\caption{\small A failure mode of the scoring method. We revisit the Branin experiment from \ref{subsec:ranking} but instead pick five observed points intentionally from a bad region. Although the search spaces including the best and worst observations overlap (left), the empirical scores suggest a clear ranking of the three search spaces, with the loose search space ranked highest. However, the predicted scores view the search spaces as nearly equivalent due to the poor GP fit.
} 
\label{fig:fail-branin}
\end{figure}%
\begin{figure}[t!]
\centering
\includegraphics[width=0.3\textwidth]{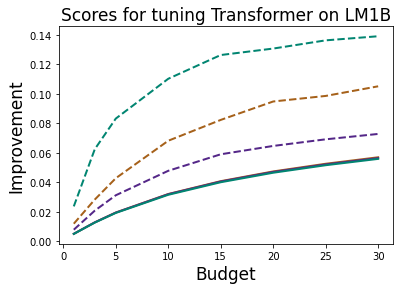}

\caption{\small 
Reproducing Fig. \ref{fig:hp-importance}, right, with a much worse GP that only observes five points, all with relatively poor performance causes the predicted scores to be nearly useless. 
}
\label{fig:fail-lm1b}
\vspace{-1em}
\end{figure}

Note that the first case, i.e., comparing two random search spaces is more challenging than the second case, comparing a random search space with the max-scoring search space. Regardless, we see in Fig. \ref{fig:prob-rank-preserve} that our scores correctly rank well-separated search spaces with high probability in both cases. Intuitively, the ranking accuracy increases as the search spaces differ more significantly.
\subsection{Pruning bad search space regions} \label{subsec:prune-bad}
We consider three problems including the Hartman function, a WideResNet on CIFAR100 and a ResNet50 on ImageNet, using budgets of $B=60,~50,~70$, respectively. For each problem, we set $b_1=b_2=B/2$ and let \texttt{sample\_observations} be random search and implemented \texttt{propose\_search\_spaces} as described in Appendix \ref{subsubsec:app-ss-random} with volume reduction rates $\rho \in \{0.1,~ 0.2,\dots,0.9\}$ with 500 search spaces per rate. We ran Alg. \ref{alg:ss-prune} for each problem for 100 rounds and report the mean and standard error (see Fig. \ref{fig:hartman-demo} and Fig. \ref{fig:baseline-vs-boosted}).
For all problems, random search in the max-scoring space $\gS^*$ outperforms random search over the broad search space $\gX$. For Hartman and CIFAR100, $\gS^*$ surpasses $\gX$ soon after we start spending $b_2$. For ImageNet, it takes longer for $\gS^*$ to win out. See \ref{subsec:diff-budget-splits} for the performance of CIFAR100 under different budget splits $B=b_1+b_2$ where we show our method is robust to the choice of split threshold. 

\vspace{-1em}

\subsection{Deciding whether to tune dropout}\label{subsec:hp-importance}

Consider the choice to tune or fix dropout on CIFAR100, ImageNet, and LM1B given $\lvert \gD \rvert = 35$ points of initial data, with various budgets 
of remaining evaluations. We score search spaces with dropout tuned ($\gX$), fixed to $r=0$ ($\gZ_1$), and fixed to $r=0.5$ ($\gZ_2$).
For CIFAR100 (Fig. \ref{fig:hp-importance}, left), both the predicted and empirical scores are nearly equal for all three search spaces, and indeed for WideResNet on CIFAR100 in our setup tuning dropout or fixing it to 0 or 0.5 makes no difference.
%
For ImageNet (Fig. \ref{fig:hp-importance}, middle), the predicted scores are moderately higher for $\gZ_1$ ($r=0$) and agree with the empirical ranking.
%
For LM1B (Fig. \ref{fig:hp-importance}, right), we see that the predicted scores successfully detect that at these budgets dropout should be fixed to zero and not tuned and that this choice matters a lot. Interestingly, the predicted scores have a similar relative difference as the empirical scores. On all three problems, the predicted scores succeed at deciding whether to tune dropout and correctly indicate when the choice matters, suggesting that search space scores can be a useful tool in the deep learning tuning workflow.

\vspace{-1em}
\subsection{Limitations \& failure modes} \label{sec:fail}

Our search space scores are only as good as the probabilistic model they depend on; if $p(y | x, \gD)$ is poorly calibrated or otherwise inaccurate they will produce nonsensical results. A variety of things can cause a poor GP fit, but we encountered issues most frequently when using small numbers of observations, especially when the observations only included relatively bad points.
See Fig. \ref{fig:fail-branin} for a demonstration on Branin. In addition to harming the GP fit, observed data with consistently large objective values results in an unrealistically large incumbent value that especially harms the b-PI-based scores.
Similar issues can infect our downstream applications. Revisiting deciding whether to tune dropout on LM1B, this time with $\lvert \gD \rvert=5$ with the best validation error a disappointing $y^+=0.769$ (see Fig. \ref{fig:fail-lm1b}). the predicted scores can no longer distinguish the search spaces. Finally, for pruning bad regions of a search space in \ref{subsec:prune-bad}, although our method is fairly robust to the value of splitting threshold (see \ref{subsec:app-ss}), extreme values (e.g. $b_1=1$ or $b_1=B-1$) make sampling in the pruned space $\gS^*$ no better than using $\gX$.

\section{Conclusions and Future Work}
 In this work, we motivated the problem of scoring search spaces conditioned on a budget, and demonstrated a simple method for computing such scores given any probabilistic model of the response surface. Our experiments demonstrate the proof of concept that useful scoring models will allow practitioners to answer tuning questions currently unaddressed in the literature, and greatly benefit their workflow. Applications of accurate search space scores will go beyond simply pruning bad search spaces in iterated studies, they have the potential to help mitigate the ML reproducibility crisis by allowing researchers to provide guidelines on how to best reproduce their results conditioned on a budget. We hope that this will enable the community to develop robust methods that can be deployed broadly and reduce effort adapting methods to new settings.
 
 Our search space scoring functions are based on a general expected utility framework and thus can be extended in a variety of ways. They are agnostic to the choice of model for $p(\mathbf{y} | \mathbf{x}, \mathcal{D})$ and alternative models such as random forests or Bayesian neural networks might, e.g.,\ scale better. We can also generalize the improvement-based utility $u(y^+,\mathbf{y})$ to  $u_g(\theta,\mathcal{D},\mathbf{x},\mathbf{y})$ where $\gD$ is the observed data and $\theta$ denotes additional parameters of the generalized utility function. For example, inspired by batch Bayesian optimization, $u_g(\theta,\mathbf{x},\mathbf{y})$ could be the amount of information gained about the location of the minimum of $f$ from observing $\mathbf{x},\mathbf{y}$ \citep{shah2015parallel}. Such a choice of utility function would remove the direct dependence of the scores on the incumbent $y^+$, but it might come with new challenges, namely search spaces that produce the most information about the location of the minimum are not necessarily the same ones that will result in better points. Information-theoretic utility functions also might be more difficult to estimate than our simple improvement-based utility functions. Finally, we could explore scores based on non-uniform sampling of $\mathbf{x}$ locations, that is to say $\mathbf{x}\sim \Delta_b(\gS,\upsilon)$ where $\upsilon$ denotes additional parameters of the sampling distribution.
 
 Moving forward, there are many potential directions to build on our search space scores. Potential extensions to our framework include: more general sampling strategies, more general budget constraints (such as time), and using a score function to propose optimal search space bounds for a given budget. Beyond extending the framework, there is more to be done on the modeling side. While our scores usually were rank preserving, they sometimes overestimated the magnitude of improvement, and performed poorly when conditioned on little data. Better model specification, e.g.\ through bespoke or data-dependent priors, could mitigate these issues. Overall, developing better budget-aware search space scores would unlock new applications for Bayesian optimization and we hope our  method serves as a useful starting point for the community to tackle this challenging problem.

\bibliography{main_paper}
\bibliographystyle{abbrvnat}
\clearpage
\appendix
\section{Appendix}\label{sec:app}
\subsection{Background: Bayesian Optimization}\label{app-gp}
Bayesian Optimization (BO) is an automated method for optimizing expensive black-box objective functions. Consider a black-box (expensive to query) objective function $f$ over domain $\mathcal{X}$ that we wish to minimize. The observed values of $f$ are potentially corrupted with noise. Given a set of observed data $\mathcal D=\{(x_j,y_j)\}_{j=1}^n$ where $y_i\sim\text{Normal}(f(x_j),\sigma_{\text{noise}}^2)$, BO builds a response surface model for $f$, typically assumed to be a Gaussian Process (GP). GPs are probabilistic non-parametric models and a popular choice for modelling smooth functions because they allow closed-form inference and can naturally accommodate prior knowledge~\citep{rasmussen2003gaussian}.

Let $x_{(1:n)}$ and  $y_{(1:n)}$ denote the observed data in $\mathcal{D}$. Given a batch of points $\mathbf{x}=\{x_i\}_{i=1}^m$, the posterior distribution over the values of $f$ at $\mathbf{x}$ is a multivariate Gaussian with mean 
\begin{equation*}
    \mu(\mathbf{x}) = \mathbf{k}(\mathbf{x},x_{(1:n)})\Big(\mathbf{K} + \sigma_{\text{noise}}^2 I\Big)^{-1} y_{(1:n)}
\end{equation*}
and covariance
\begin{equation*}
    \Sigma(\mathbf{x})=\mathbf{k}(\mathbf{x},\mathbf{x}) - \mathbf{k}(\mathbf{x},x_{(1:n)})\Big(\mathbf{K} + \sigma_{\text{noise}}^2 I\Big)^{-1}\mathbf{k}(x_{(1:n)},\mathbf{x}),
\end{equation*}
where $\mathbf{K}=[k(x_j,x_j')]_{j,j'=1}^{n}$, $\mathbf{k}(\mathbf{x},\mathbf{x})=[k(x_i, x_{i'})]_{i,i'=1}^m$, and
$\mathbf{k}(\mathbf{x},x_{(1:n)})=[k(x_i,x_j)]_{i,j=1}^{m,n}$.
Bayesian optimization selects the $x$-location of the next batch of points to query $f$ by maximizing an acquisition function over a sample batch of points $\mathbf{x}$.
The value of the acquisition function for a batch $\mathbf{x}$ is the expected \emph{utility} of the batch, where the expectation is taken with respect to the objective values at $\mathbf{x}$, that is $\mathbf{y}|\mathbf{x},\mathcal{D}$. 
Two popular choices of utility functions are the indicator and the magnitude of the improvement over the best value of $\mathbf{y}$, also known as the incumbent value \citep{movckus1975bayesian, jones1998efficient}. See \citep{frazier2018tutorial} for different choices of acquisition functions and details of BO algorithm.

\subsection{Experimental Setup}\label{app:more-plots}
In this section, we describe the details of the specific design and modeling choices we made for our experiments in Section \ref{sec:experiments}. In Section \ref{app:gp-details}, we explain the GP implementation details, including the choice of the mean and the covariance functions, the hyperpriors, and the reparameterization we applied to the GP hyperparameters. Next, in Section \ref{subsec:app-ss}, we discuss the procedures we followed to generate search spaces that we score in Section \ref{subsec:ranking} and Section \ref{subsec:prune-bad}, although many other alternative procedures are possible. We make no claim to have found any sort of optimal procedure. Finally, in Section \ref{app:ss-details}, we include the specific bounds of the search spaces we scored in Section \ref{subsec:exp-details} for tuning a WideResNet on CIFAR100 and for tuning a ResNet50 on ImageNet.
\subsubsection{GP implementation details}\label{app:gp-details}
In our experiments, we used GPs with constant, zero mean and an ARD Mat{\'e}rn-5/2 kernel. For the purposes of optimization, we reparameterized amplitude, lengthscales and noise variance using a softplus function. We place standard zero-mean, unit variance log-Normal priors on the amplitude, and inverse-lengthscales, and a zero-mean $0.1$-variance normal on the noise term. The GP hyperparameters are fitted by optimizing the log of the marginal likelihood function via 3000 steps of LBFGS \citep{liu1989limited}. 

\subsubsection{Search space generation}\label{subsec:app-ss}
Throughout our experiments, we followed a simple volume-constrained procedure to generate search spaces, depending on the situation, either centering them on a particular point or placing them randomly with a base search space. Given a volume reduction rate $\rho \in (0, 1]$ and a $d$-dimensional base hyperrectangular search space $\mathcal{X}$, we consider two scenarios. First, building a random search space in $\mathcal{X}$ with volume $\rho$ times the volume of $\mathcal{X}$ and second, building a search space centered at a point $x \in \mathcal{X}$ with approximate volume $\rho$ times the volume of $\mathcal{X}$. In both cases the generated search space is a subset of the base search space $\gX$.

\paragraph{Random search space generation}\label{subsubsec:app-ss-random}
Let $\mathcal{S}$ be the search space we want to generate where $\mathcal{S}\subset \mathcal{X}$. We denote the i$th$ dimension of $\mathcal{X}$ and $\mathcal{S}$ with $\mathcal{X}^i$ and $\mathcal{S}^i$ accordingly. For each dimension $i,~i=1,\dots,d$, we compute the dimension-wise reduction rate as $\rho_i = \rho^{1/d}$. The length of $\mathcal{S}^i$ is denoted by $l_i=\rho_i \times \big(\text{max}(\mathcal{X}^i)-\text{min}(\mathcal{X}^i)\big)$. We generate a random value in $\big[\text{min}\big(\mathcal{X}^i\big),~l_i\big]$ and set it as $\text{min}\big(\mathcal{S}^i\big)$. We have $\text{max}\big(\mathcal{S}^i\big)=\text{min}\big(\mathcal{S}^i\big)+l_i$ and $\mathcal{S}^i=\big[\text{min}\big(\mathcal{S}^i\big),~\text{max}\big(\mathcal{S}^i\big)\big]$ and we repeat this procedure for each $i$ and stack the intervals $\gS^i$ to generate $\mathcal{S}$.

\paragraph{Search space generation centered at a point}\label{subsubsec:app-ss-around-point}
Let $x\in\mathcal{X}$ denote the point we want to generate search space $\mathcal{S}\subset \gX$ around it. For each dimension $i$, we calculate the interval length $l_i$ as explained in Section \ref{subsubsec:app-ss-random}. We set $\text{min}\big( \mathcal{S}^i \big) = \text{max}\big(\text{min}\big(\mathcal{X}^i\big),x^i-\frac{1}{2}l_i$\big) and $\text{max}\big( \mathcal{S}^i \big) = \text{min}\big(\text{max}\big(\mathcal{X}^i\big),x^i+\frac{1}{2}l_i$\big). As before, $\mathcal{S}^i=\big[\text{min}\big(\mathcal{S}^i\big),~\text{max}\big(\mathcal{S}^i\big)\big]$ and we repeat this procedure for each $i$ and stack the intervals $\gS^i$ which generates $\mathcal{S}$. 

Note that because of the clipping of the bounds, the volume constraint might not be precisely met given the location of $x$, for example when $x$ is relatively close to the boundaries of $\gX$. Although this might affect the volume of some of the generated search spaces in our experiments, it does not affect our experimental conclusions. Hence, with a slight misuse of terminology, we still refer to a generated search space by its \emph{targeted} volume reduction rate.
\subsubsection{Tuning details \& Search Spaces}\label{app:ss-details}
Here we specify the search spaces we have scored and ranked in the experiments Sections \ref{subsec:ranking}, \ref{subsec:prune-bad} and \ref{subsec:hp-importance}. For all tuning problems, including tuning a WideResNet on CIFAR100, tuning a ResNet50 on ImageNet, and tuning a transformer on LM1B, we have used the same broad, seven-dimensional, base hyperparameter search space. This search space includes the base learning rate $\eta$, momentum $\beta$, learning rate decay power $p$, decay step factor $\tau$, dropout rate $r$, $l_2$ decay factor $\lambda$, and label smoothing $\gamma$. See Table \ref{tab:loose-base-ss} for specific lower and upper bounds on each hyperparameter. Note that following~\citet{NadoEtAl2021_largeBatchReality}, we tune one minus momentum instead of momentum. Additionally, we tune the base learning rate, one minus momentum and $l_2$ decay factor $\lambda$ over a $\log_{10}$ scale. For each hyperparameter except for the dropout rate that was sampled from a discrete set, the reparameterized hyperparameter locations were sampled uniformly at random within the scaled specified bounds. The dropout locations were sampled uniformly at random over $\{0,~ 0.1,~0.2,\dots, ~0.7,~ 0.8\}$. 
All the concrete search spaces we used in experiments used the same hyperparameter scaling and reparameterization choices.

\paragraph{Tuning a WideResNet on CIFAR100}
In Section \ref{subsec:ranking}, we score three search spaces for tuning a WideResNet on CIFAR100, including the base loose search space, defined in Table \ref{tab:loose-base-ss}, and two volume-constrained search spaces generated following the second search space generation strategy in Section \ref{subsubsec:app-ss-random} with volume reduce rates $\rho\in\{0.25,~0.5\}$. Both reduced search spaces are centered around the observed hyperparameter point in $\gD$ with the median validation error in $\gD$. See bounds of these search spaces in Table \ref{tab:cifar100-.25} and Table \ref{tab:cifar100-.5}. In Sections \ref{subsec:prune-bad} and \ref{subsec:hp-importance}, the base loose search space is as defined in Table \ref{tab:loose-base-ss}. In Section \ref{subsec:hp-importance}, in addition to the base loose search space, we consider two other search spaces. For each search space, all the hyperparameters except for the dropout rate $r$ have similar bounds, scaling and reparameterization as in the base loose search space stated in Table \ref{tab:loose-base-ss}. For the dropout rate, we fix it to zero for one search space and to $0.5$ for another one, that is, following the notation in Section \ref{subsec:hp-importance}, $r \in \{0,~0.5\}$. 

\paragraph{Tuning a ResNet50 on ImageNet}
In Section \ref{subsec:ranking}, we score three search spaces for tuning a ResNet50 on ImageNet, including the base loose search space, defined in Table~\ref{tab:loose-base-ss}, and two volume-constrained search spaces generated following the second search space generation strategy in Section \ref{subsubsec:app-ss-random} with volume reduction factors $\rho\in\{0.25,~0.5\}$. Both reduced search spaces are centered around the observed hyperparameter point in $\gD$ with the best (minimum) validation error amongst all validation errors in $\gD$. Bounds of these search spaces can be found in Table~\ref{tab:imagenet-.25} and Table~\ref{tab:imagenet-.5}. In Sections \ref{subsec:prune-bad} and \ref{subsec:hp-importance}, the base loose search space is as defined in Table~\ref{tab:loose-base-ss}. In \ref{subsec:hp-importance}, in addition to the base loose search space, we consider two other search spaces. For each search space, all the hyperparameters except for the dropout rate $r$ have similar bounds, scaling and reparameterization as in the base loose search space stated in Table \ref{tab:loose-base-ss}. For the dropout rate, we fix it to zero for one search space and to $0.5$ for another one, that is, following the notation in Section \ref{subsec:hp-importance}, $r \in \{0,~0.5\}$. 

\paragraph{Tuning a Transformer on LM1B}
 In Section \ref{subsec:hp-importance},  we score three search spaces for tuning a Transformer on LM1B, including the base loose search space, defined in Table \ref{tab:loose-base-ss}, and two other search spaces. For each search space, all the hyperparameters except for the dropout rate $r$ have similar bounds, scaling and reparameterization as in the base loose search space. For the dropout rate, we fix it to zero for one search space and to $0.5$ for another one, that is, following the notation in Section \ref{subsec:hp-importance}, $rb\in \{0,~0.5\}$.



\begin{table}[!htbp]
   \caption{\small The base loose search space including the base learning rate $\eta$, momentum $\beta$, power $p$, decay step factor $\tau$, dropout rate $r$, $l_2$ decay factor $\lambda$ and label smoothing $\gamma$. }
  
   \vspace{-.4em}
    \begin{center}
    \scalebox{1}{
        \begin{tabular}{ccccccc}
        \toprule
       \textsc{Hyperparameter}   & \textsc{Log$_{10}$}  & \textsc{Min} & \textsc{Max}   \\\midrule
         $\eta$ &   \checkmark    & -5&1  \\ 
         $1-\beta$ &  \checkmark    &-3 & 0 \\ 
         $p$ &    -  &0.1 &2  \\ 
         $\tau$  & -  & 0.01  &0.99  \\ 
         $r$ &   -   & 0.1&0.8  \\ 
          $\lambda$ &  \checkmark     &-6 &-0.69  \\ 
        $\gamma$  &  -    & 0&  0.4\\ 
                                               \bottomrule
        \end{tabular}
    }
    \end{center}
    \vspace{.5em}
    \label{tab:loose-base-ss}
\end{table}

\begin{table}[!htbp]
   \caption{\small The search space with approximately $0.25$ of the volume of the base search space for CIFAR100 experiment in Section \ref{subsec:ranking}. }
  
   \vspace{-.3em}
   \label{tab:summary}
    \begin{center}
    \scalebox{1}{
        \begin{tabular}{ccccccc}
            \toprule
           \textsc{Hyperparameter}   & \textsc{Log$_{10}$}  & \textsc{Min} & \textsc{Max}   \\\midrule
             $\eta$ &   \checkmark    & -5&-1.18  \\ 
             $1-\beta$ &  \checkmark    &-3 & -0.96 \\ 
             $p$ &    -  &0.98 &2  \\ 
             $\tau$  & -  & 0.01  &0.73  \\ 
             $r$ &   -   & 0.1&0.53  \\ 
              $\lambda$ &  \checkmark     &-6 &-3.33  \\ 
            $\gamma$  &  -    & 0&  0.3\\ 
                                                   \bottomrule
            \end{tabular}
    }
    \end{center}
    \label{tab:cifar100-.25}
\end{table}
\begin{table}[t!]
   \caption{\small The search space with approximately $0.5$ of the volume of the base search space for CIFAR100 experiment in Section \ref{subsec:ranking}. }
  
   \vspace{-.4em}
   \label{tab:summary}
    \begin{center}
    \scalebox{1}{
 \begin{tabular}{cccc}
        \toprule
       \textsc{Hyperparameter}   & \textsc{Log$_{10}$}  & \textsc{Min} & \textsc{Max}   \\\midrule
         $\eta$ &   \checkmark    & -5&-0.92  \\ 
         $1-\beta$ &  \checkmark    &-3 & -0.83 \\ 
         $p$ &    -  &0.9 &2  \\ 
         $\tau$  & -  & 0.01  &0.78  \\ 
         $r$ &   -   & 0.1&0.56  \\ 
          $\lambda$ &  \checkmark     &-6 &-2.99  \\ 
        $\gamma$  &  -    & 0&  0.2\\ 
                                               \bottomrule
        \end{tabular}
    }
    \end{center}
    \vspace{.5em}
    \label{tab:cifar100-.5}
\end{table}

\begin{table}[!htbp]
     \caption{\small The search space with approximately $0.25$ of the volume of the base search space for ImageNet experiment in Section \ref{subsec:ranking}. }
  
   \vspace{-.3em}
    \begin{center}
    \scalebox{1}{
            \begin{tabular}{ccccccc}
        \toprule
       \textsc{Hyperparameter}   & \textsc{Log$_{10}$}  & \textsc{Min} & \textsc{Max}   \\\midrule
         $\eta$ &   \checkmark    & -1.81&1  \\ 
         $1-\beta$ &  \checkmark    &-2.58 & -0.11 \\ 
         $p$ &    -  &0.1 &1.37  \\ 
         $\tau$  & -  & 0.42  &0.99  \\ 
         $r$ &   -   & 0&0.53  \\ 
          $\lambda$ &  \checkmark     &-6 &-3.65  \\ 
        $\gamma$  &  -    & 0.11&  0.4\\ 
                                               \bottomrule
        \end{tabular}
    }
    \end{center}
    \vspace{.5em}
     \label{tab:imagenet-.25}
\end{table}

\begin{table}[!htbp]
     \caption{\small The search space with approximately $0.5$ of the volume of the base search space for ImageNet experiment in Section \ref{subsec:ranking}. }
  
   \vspace{-.3em}
    \begin{center}
    \scalebox{1}{
        \begin{tabular}{ccccccc}
        \toprule
       \textsc{Hyperparameter}   & \textsc{Log$_{10}$}  & \textsc{Min} & \textsc{Max}   \\\midrule
         $\eta$ &   \checkmark    & -2.06&1  \\ 
         $1-\beta$ &  \checkmark    &-2.7 & 0 \\ 
         $p$ &    -  &0.1 &1.45  \\ 
         $\tau$  & -  & 0.37  &0.99  \\ 
         $r$ &   -   & 0&0.56  \\ 
          $\lambda$ &  \checkmark     &-6 &-3.42  \\ 
        $\gamma$  &  -    & 0.09&  0.4\\ 
                                               \bottomrule
        \end{tabular}
    }
    \end{center}
     \label{tab:imagenet-.5}
\end{table}
 \paragraph{Decay Schedule for Tuning}
For all the tuning problems, we used the following polynomial decay schedule for the learning rate
\[
\eta_t = \frac{\eta}{1000}+\big(\eta - \frac{\eta}{1000}\big)\Big(1-\frac{\text{min}(t,\tau T)}{\tau T}\Big)^p,
\]
where $t$ denotes the current training step and $T$ is the total number of training steps.
\vspace{10em}
\subsection{Additional Results \& Sensitivity Analysis for pruning bad regions of the search space}\label{app:subsec:sa}
Here, we report some additional results that were omitted from the main manuscript to save space. In Section \ref{app:subsec:hartman} we report the result of pruning bad regions of the search space for the six-dimensional Hartman function. Next, in Section \ref{subsec:diff-scores} we report on the performance of variations of the scores including median-b-EI, mean-b-PI, and median-b-PI and compare them to our baseline score mean-b-EI for tuning a WideResNet on CIFAR100 and for minimizing Hartman; ultimately the different variation are quite correlated. Finally, in Section \ref{subsec:diff-budget-splits}, we analyze the sensitivity of search space pruning results for tuning WideResNet on CIFAR100 to the budget split $B=b_1+b_2$. We show that regardless of the specific budget split (within reason), our strategy of pruning bad regions a base loose search space and running random search on the pruned space consistently outperforms running random search in the base search space.


\subsubsection{Pruning bad search space regions for Hartman}\label{app:subsec:hartman}
\begin{figure}[htp!]
  \centering
\includegraphics[width=0.3\textwidth]{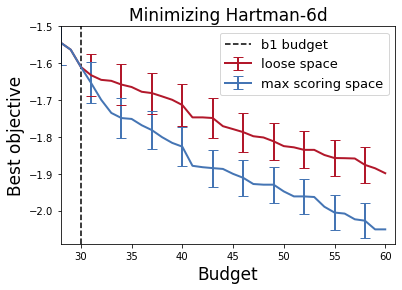}

\caption{\small We show that minimizing the six-dimensional synthetic Hartman function over max-scoring search space versus the enclosing loose search space can achieve better results, faster (see Section \ref{subsec:prune-bad} for more details).
}
\label{fig:hartman-demo}
\end{figure}

Here, we report on the effect of pruning bad regions of the base loose search space, i.e. $[0,~1]^6$ for minimizing the Hartman function. Following Section \ref{subsubsec:app-ss-random}, we have generated 500 random search spaces for each volume reduction rate $\rho \in \{0.1,~ 0.2,\dots,0.9\}$. After spending $b_1$ budget uniformly at random within the base search space, we score all the generated search spaces and the base search space using the model trained on the $b_1$ points and spend the remaining $b_2$ in the max-scoring search space. We compare the performance of this procedure to the baseline, that is spending the entire budget $B=b_1+b_2$ within the base search space. We repeat this for $100$ random initialization and report the mean performance with standard errors in Fig. \ref{fig:hartman-demo}. As portrayed in Fig. \ref{fig:hartman-demo}, pruning bad regions of the base search space improves the performance of random search for minimizing the Hartman function. 
\begin{figure*}[t!]
    \centering
    \includegraphics[width=0.8\textwidth]{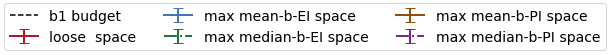}\quad\quad\quad\quad\quad\quad\quad\quad\quad\quad
    \includegraphics[width=0.3\textwidth]{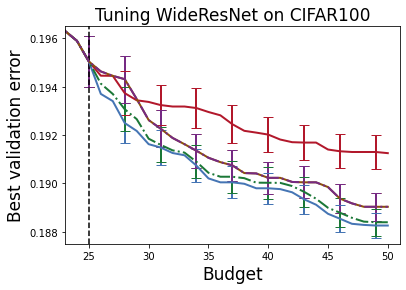}
    \includegraphics[width=0.3\textwidth]{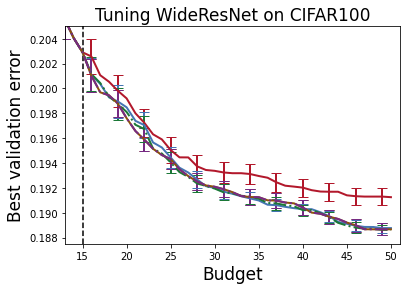}
    \includegraphics[width=0.3\textwidth]{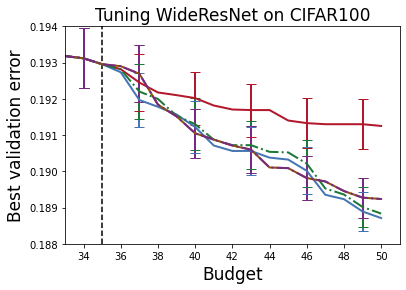}
    \caption{Comparison of the effect of different splitting budget thresholds $B=b_1+b_2$ in Alg. \ref{alg:ss-prune} on tuning a WideResNet on CIFAR100. Interestingly, all budget split choices result in improvement upon the baseline. For all splitting choices, the larger the budget, the more significant the improvement becomes.}
    \label{fig:budget-splits}
\end{figure*}

\subsubsection{Performance of different score variants}\label{subsec:diff-scores}
\begin{figure}[htp!]
  \centering
\includegraphics[width=0.28\textwidth]{figs/finalized/wide_resnet_cifar100_all_scores.png}\quad\quad\quad\quad\quad\quad\quad\quad
\includegraphics[width=0.28\textwidth]{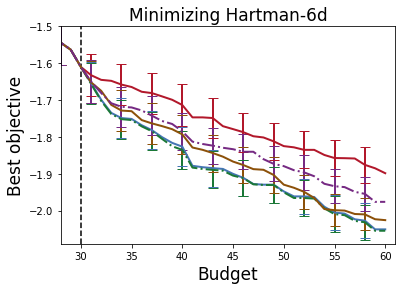}

\caption{\small Comparison of random search within a max-scoring (according to different score variants) search space and the enclosing, loose search space. See Fig. \ref{fig:budget-splits} for the legend. Mean-b-EI and median-b-EI show almost identical performance for Hartman and very similar performance for CIFAR100. For Hartman, median-b-PI outperforms mean-b-PI while both scores perform identically for CIFAR100. For both Hartman and CIFAR100, the b-EI scores slightly outperform the b-PI scores. Overall, all the score variations outperform the baseline which shows that pruning the bad regions of the loose search space can improve the results, regardless of the pruning method details.}
\label{fig:diff-scores}
\end{figure}

Here, we compare the performance of different score variants including mean-b-EI, median-b-EI, mean-b-PI, median-b-PI when used for pruning bad regions of the base loose search space. We consider two problems including tuning a WideResNet on CIFAR100 and  minimizing the six-dimensional Hartman function. See Section \ref{subsec:prune-bad} for more details. As mentioned before, all the scores perform similarly. Naturally, given an expected utility (e.g. either b-EI or b-PI), the mean and the median scores are often strongly correlated (mean-b-EI and median-b-EI, or, similarly, mean-b-PI and median-b-PI). Given a centrality measure over the $x$-locations (e.g. mean or median), the scores are still correlated but the b-EI vs b-PI distinction seems to matter a bit more than the mean vs median distinction. Overall, pruning bad regions of the base loose search space using any of the scores and running random search over the max-scoring search space results in improvement upon the baseline of random search on the enclosing base search space. See Fig. \ref{fig:budget-splits} and Fig. \ref{fig:diff-scores} for an illustration.

\subsubsection{The effect of budget splitting threshold for pruning bad search space regions}\label{subsec:diff-budget-splits} 

Finally, we show that for tuning a WideResNet on CIFAR100, our score-based pruning method is robust not only to the particular score function variant, but to the choice of how to split the total budget $B=b_1+b_2$ between phases. Recall that Alg.~\ref{alg:ss-prune} splits the entire available budget $B=b_1+b_2$, spends the initial budget $b_1$ in the loose search space, and the remaining budget $b_2$ on the max-scoring search space. As shown in Fig. \ref{fig:budget-splits}, our scoring method for the CIFAR100 tuning problem is relatively insensitive to the choice of splitting threshold in the sense that it can consistently achieve better results compared to the baseline of expending the entire budget in the enclosing, loose search space. We consider $B=50$ and report the results over $b_1 \in \{15,~25,~35\}$, trying all four score variants (mean-b-EI, median-b-E, mean-b-PI and median-b-PI). Consistently, for all the scores across all the splitting thresholds, the score-guided, two-phase tuning outperforms tuning over the loose search space. This result shows the robustness of these scoring functions and their consistent capability to improve upon the baseline.

\begin{figure}[t!]
  \centering
\includegraphics[width=0.4\textwidth]{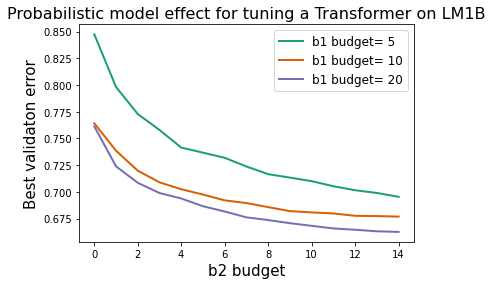}
\caption{\small Comparing the impact of budget $b_1$ on the fitted distribution $p(y | x, \gD)$ and pruning bad regions of the search space. Larger $b_1$ values result in better fitted models, hence more promising max-scoring search spaces.}
\label{fig:effect_of_prob_model_on_sss}
\end{figure}
Additionally, we investigated the effect the quality of $p(y | x, \gD)$ has on the search space pruning results for tuning a Transformer model on LM1B. By varying the amount of training data for $p(y | x, \gD)$ we can explore probabilistic models of different qualities. Specifically, we vary $b_1$ while keeping $b_2$ fixed in our search space pruning experiment. As shown in Fig. \ref{fig:effect_of_prob_model_on_sss}, the larger $b_1$, the better the fitted distribution $p(y | x, \mathcal{D})$ is and the better the max-scoring search space will perform after the additional $b_2$ evaluations. 
\end{document}